\documentclass[letterpaper]{article} 
\usepackage{aaai2027}  
\usepackage[hyphens]{url}  
\usepackage{graphicx} 
\usepackage{natbib}  
\usepackage{caption} 
\usepackage{algorithm}
\usepackage{algorithmic}
\usepackage{booktabs}
\usepackage{amsmath,amssymb}
\usepackage{multirow}
\usepackage[export]{adjustbox}
\usepackage{subcaption}
\usepackage{tikz}
\usetikzlibrary{arrows.meta,positioning,shapes.geometric,fit,backgrounds}

\title{An Emerging Retail Portfolio Management Application:\\
Personalized, Tax-Aware Reinforcement Learning with Natural Language Goals}

\author{
    Ramin Pishehvar\thanks{Patent Pending. U.S. Provisional Patent Application No. 64/101,198, filed June 29, 2026; and U.S. Provisional Patent Application No. 64/117,071, filed July 22, 2026.}
}
\affiliations{
    rpichevar@gmail.com
}

\begin{document}

\maketitle

\begin{abstract}
Retail investors lack access to the kind of personalized, tax-aware
portfolio management that institutional clients take for granted --
existing robo-advisors use static, rule-based allocation, and
institutional-grade systems require account minimums and technology
stacks unavailable to individual investors. We present a fully built,
integration-tested application that closes this gap: a FastAPI backend
and web dashboard
that let a user describe an investment goal in plain language (e.g.\
``I want steady growth but need to sell some shares next month for a
down payment''), routes that goal to one of six investment mandates,
and produces a live, broker-integrated portfolio recommendation from a
three-phase reinforcement learning system -- a self-supervised
cross-asset encoder, a Mixture-of-Experts (MoE) allocation policy with
a learned intent router, and a lightweight LoRA adapter that
personalizes recommendations from an individual's revealed brokerage
behavior without retraining the shared model. The system is
functionally complete and integration-tested end-to-end against a live
brokerage API (Alpaca, paper-trading mode), including multi-user
authentication, a trust-first preview-before-apply confirmation flow,
daily email digests, and an auditable action-integrity chain, but has
not yet been opened to real end-users; we report this honestly as an
emerging, pre-deployment application with a concrete path to full
deployment, alongside 14-day walk-forward backtests (bootstrapped
confidence intervals included) as preliminary, pre-deployment
validation rather than production performance. We also report several
practical engineering lessons -- silently-inactive integration paths,
hanging third-party API calls, and the value of end-to-end empirical
verification over trusting checkpoint metadata -- that we believe
generalize to other applied RL systems built on external, live data
sources.
\end{abstract}

\section{Problem and Motivation}
Robo-advisors (Betterment, Wealthfront, Schwab Intelligent Portfolios)
allocate retail accounts using static, rule-based glide paths tied to
age and a coarse risk questionnaire -- the same allocation logic
applies whether a user's actual goal is a 30-day trade or a 20-year
retirement horizon. Institutional portfolio management -- factor
models, tax-lot optimization, personalized mandates -- exists, but
requires account minimums, dedicated relationship managers, and
technology budgets that put it out of reach for most individual
investors. The result is a widening gap: sophisticated, learned
portfolio management for institutions, and simple heuristics for
everyone else.

We built and are integration-testing an application intended to close
this gap directly: a system that takes a plain-language description of
an investor's actual goal, maps it onto one of six concrete investment
mandates (short-term alpha, long-term growth, capital preservation,
income/tax-loss harvesting, and two personalization-relevant
variants), and produces a live, executable portfolio recommendation
that a user reviews and confirms before anything reaches their
brokerage account.

\begin{figure*}[t]
\centering
\begin{tikzpicture}[
  font=\scriptsize,
  box/.style={draw, rounded corners=3pt, minimum width=2.2cm,
              minimum height=0.75cm, align=center, fill=#1!12},
  sbox/.style={draw, rounded corners=2pt, minimum width=1.8cm,
               minimum height=0.6cm, align=center, fill=#1!8,
               font=\scriptsize},
  arrow/.style={-{Stealth[length=5pt]}, thick},
  darrow/.style={-{Stealth[length=4pt]}, thick, dashed, gray},
  label/.style={font=\scriptsize\itshape, text=gray, align=center}
]

\node[box=blue]  (p1)   at (0,0)     {Phase 1\\SSL Pretraining};
\node[box=blue]  (enc)  at (2.8,0)   {CrossAsset\\Encoder};
\node[sbox=blue] (chr)  at (1.8,-1.4) {+ Chronos\\(frozen)};
\node[sbox=blue] (news) at (3.8,-1.4) {News/Events\\(optional)};
\node[sbox=blue] (meta) at (2.8,-2.6) {50-dim\\Metadata};
\draw[arrow]  (p1)  -- (enc);
\draw[darrow] (chr)  -- (enc);
\draw[darrow] (news) -- (enc);
\draw[darrow] (meta) -- (enc);

\node[box=orange]  (p2)  at (6.0,0)   {Phase 2\\PPO Fine-tune};
\node[box=orange]  (pac) at (9.0,0)   {Portfolio\\ActorCritic};
\node[sbox=orange] (rwd) at (6.0,-1.4) {Shaped\\Reward};
\node[sbox=orange] (obj) at (6.0,-2.5) {6 Objectives\\per episode};
\draw[arrow]  (enc) -- (p2);
\draw[arrow]  (p2)  -- (pac);
\draw[darrow] (rwd) -- (p2);
\draw[darrow] (obj) -- (rwd);

\node[box=green!60!black]  (p3)   at (12.0,0)  {Phase 3\\Personalise};
\node[box=green!60!black]  (pers) at (15.0,0)  {Persona\\+ LoRA};
\node[sbox=green!60!black] (txn)  at (11.0,-1.4) {Brokerage\\transactions};
\node[sbox=green!60!black] (nlp)  at (13.0,-1.4) {NL intent\\parser};
\draw[arrow]  (pac)  -- (p3);
\draw[arrow]  (p3)   -- (pers);
\draw[darrow] (txn)  -- (p3);
\draw[darrow] (nlp)  -- (p3);

\node[box=purple] (out) at (7.5,-4.0)
  {Live recommendations: BUY $\mid$ HOLD $\mid$ SELL + weights};
\draw[arrow] (pers.south) |- (out.east);

\node[label] at (9.0,-1.2) {ticker-identity-free\\any $N$ at inference};

\end{tikzpicture}
\caption{Three-phase pipeline. Solid arrows = training flow;
dashed = conditioning inputs (Chronos, metadata, news/events, objectives, NL parser).
Each phase is independently resumable from checkpoints.}
\label{fig:pipeline}
\end{figure*}
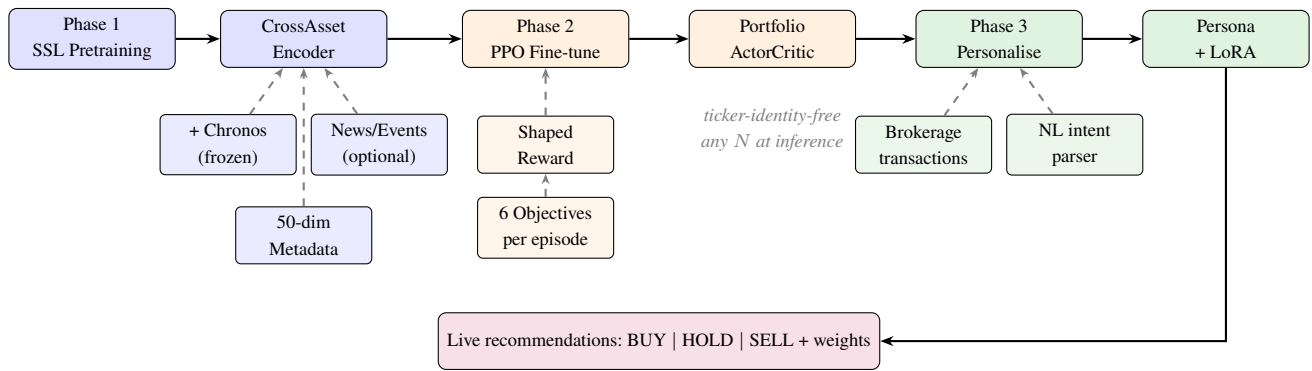
\section{System Overview}
The underlying model~\cite{pishehvar2026threephase} is trained in three
phases (Figure~\ref{fig:pipeline}); condensed architectural detail, loss
functions, and hyperparameters are in the supplementary technical
appendix.

\textbf{Phase 1 -- Cross-asset representation learning.} A
self-supervised encoder learns per-ticker representations from OHLCV
history and a 50-dimensional metadata vector (sector, market-cap
bucket, momentum/volatility statistics), making the system
ticker-identity-free: it generalizes to any publicly traded asset at
inference without retraining. Optional parallel branches fuse a frozen
time-series foundation model (Chronos-T5) and a news/event
cross-attention mechanism into the same representation. The single
most consequential training finding was \emph{representation
collapse}: without an explicit inter-ticker contrastive loss, all
tickers converged to near-identical embeddings (mean cosine similarity
0.96) and the allocation head output uniform $1/N$ weights; the
contrastive objective reduces similarity to 0.24 and restores
differentiated allocation (see the supplementary appendix).

\textbf{Phase 2 -- Mixture-of-Experts portfolio policy.} Four
specialist PPO-trained expert heads (momentum, growth, defensive,
tax-aware) are blended by a learned intent router conditioned on the
active investment mandate, so a single policy serves all six mandates
without gradient interference between them. Training the experts
requires a staged curriculum (one expert per stage, others frozen)
followed by \emph{expert grafting} -- re-inserting the best per-expert
checkpoints under the jointly-trained router -- to avoid the gradient
interference that otherwise erases specialist behavior
(see the supplementary appendix).

\textbf{Phase 3 -- Personalization.} A 76-parameter LoRA adapter shifts
policy outputs based on an individual's revealed brokerage behavior
(trade frequency, realized holding periods, tax bracket) without
retraining the shared encoder or expert heads -- personalization for a
new user is a lightweight per-user artifact ($\approx$1\,KB), not a
new model.

\section{The Application}
\label{sec:app}
This section is the paper's core contribution: a description of the
application actually built and integration-tested end-to-end -- the
server, the data layer, the brokerage abstraction, the dashboard, and
the operational controls around them. Everything described here runs;
what it has not yet done is serve real end-users trading real capital,
a distinction we keep explicit throughout rather than blurring.

\subsection{Serving Stack and Hosting}
\begin{figure*}[t]
\centering
\includegraphics[width=0.85\textwidth]{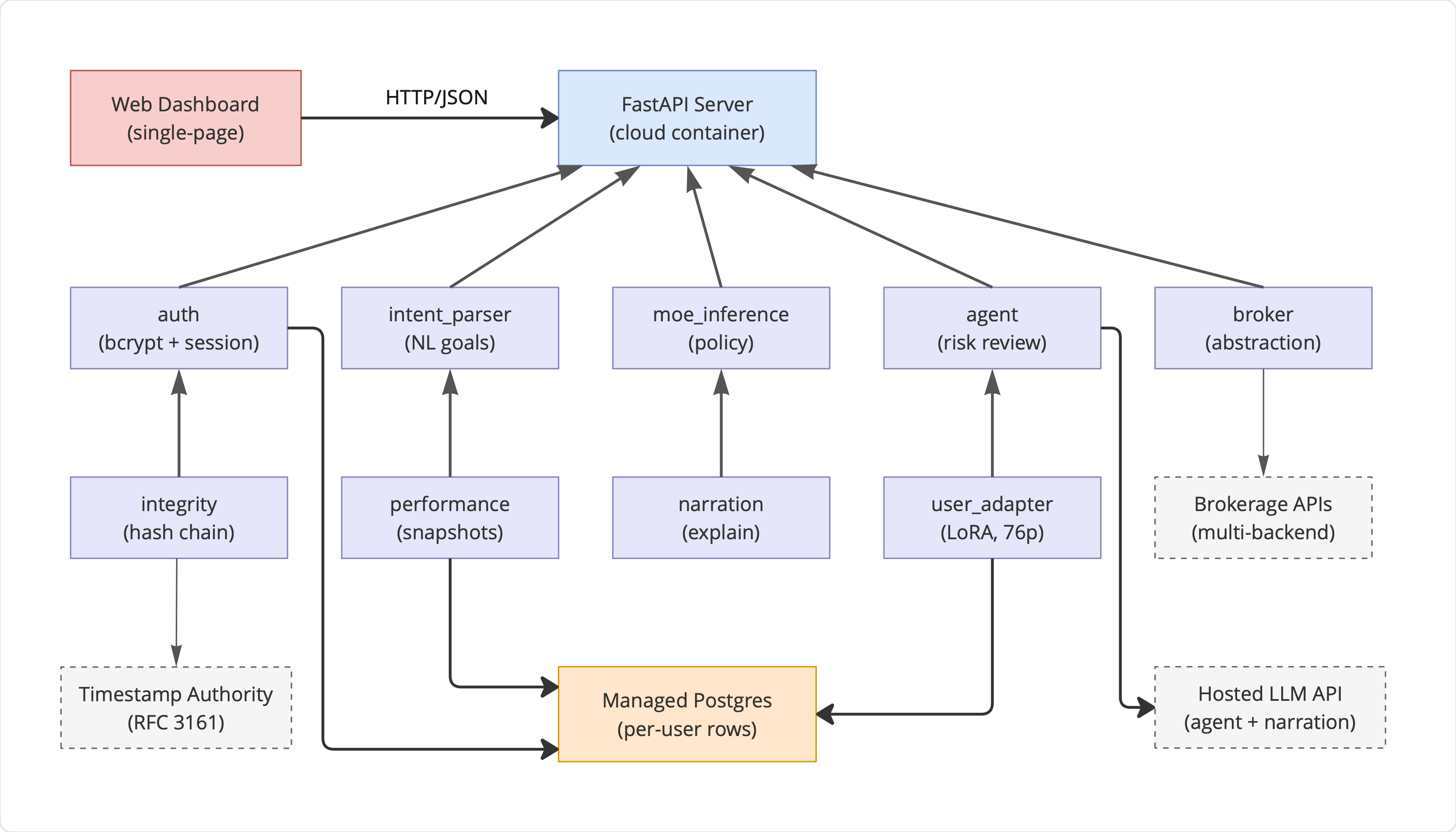}
\caption{System architecture. Solid arrows are request/data flow;
dashed arrows are internal calls or calls to a hosted LLM API. The
FastAPI server dispatches to independent modules, each named after its
source file; only \texttt{auth}, \texttt{integrity},
\texttt{performance}, and \texttt{user\_adapter} touch the database
directly, and the trained policy itself never does -- personalization
and history live in per-user rows, not in the model. }
\label{fig:architecture}
\end{figure*}

Figure~\ref{fig:architecture} shows the system architecture described
in this section and the four that follow: a single FastAPI (Python)
service serving both the JSON API and the dashboard, deployed as a
container with continuous deployment (every push rebuilds and
redeploys) and all environment-specific behavior -- broker backend,
inference engine, credentials, safety limits -- injected through
environment variables. Inference runs on CPU: the trained policy
($\sim$2M-parameter encoder plus four $\sim$200K-parameter expert heads
and a router) is small enough that a recommendation pass over a
10-ticker portfolio completes in under a second without a GPU, an
explicit goal for a retail-facing rather than institutional
application. One consequence (see Lessons Learned) is that the
deployment image pins the CPU-only PyTorch wheel; the default CUDA
build exceeded the image budget for nothing the serving path uses.

\subsection{Data Layer and Multi-User State}
Per-user state lives in a Supabase-hosted PostgreSQL database:
user accounts (username plus a bcrypt-hashed PIN, with short-lived
bearer session tokens), isolated per-user broker-connection state,
portfolio snapshots, FIFO tax-lot records for every position, pending
and confirmed recommendations, an independent per-user integrity chain
(below), and the per-user 76-parameter personalization adapters.
Keeping tax lots as first-class rows (acquisition date, cost basis,
quantity per lot) rather than aggregate positions is what allows the
application to compute the holding-period and after-tax consequences of
a proposed action at recommendation time (Tax-Lot-Aware Recommendations, below). The
model itself is user-agnostic and stateless; everything user-specific
is a database artifact, so adding a user adds rows, not model copies.

\subsection{Brokerage Abstraction}
Broker access goes through a pluggable abstraction with interchangeable
backends selected by configuration: (i) Alpaca in paper-trading mode,
used for end-to-end integration testing against a real brokerage API
surface (live price refresh, order placement, position
reconciliation); (ii) a SnapTrade OAuth integration that lets a user
connect an existing retail brokerage account (E*TRADE, Schwab,
Fidelity, and 50+ others) through a one-click flow on the dashboard;
(iii) an Interactive Brokers backend; and (iv) a deterministic mock
broker with a synthetic portfolio, used for demos and automated tests
with no external dependency. Real-money trading is gated behind a
configuration flag we have not yet enabled; every backend subclasses a
common base that enforces the same pre-trade safety checks
(Trust and Safety Controls, below) before any order is constructed.

\subsection{Dashboard}
\begin{figure*}[t]
\centering
\newlength{\figH}\setlength{\figH}{3.35in}%
\begin{minipage}[t]{2.95in}
\centering
\includegraphics[height=\figH,valign=t]{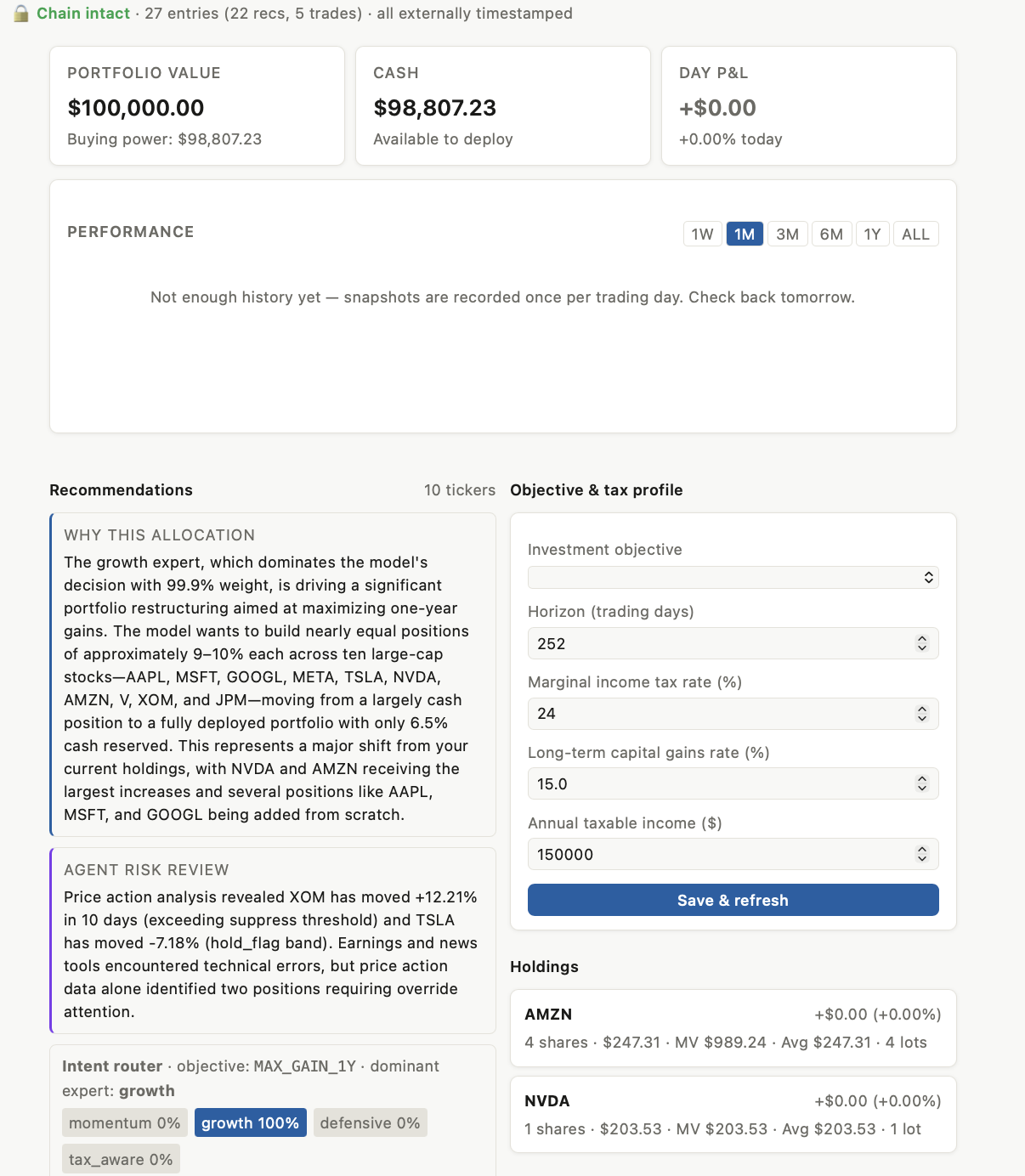}
\subcaption{Portfolio overview: integrity-chain status, natural-language
allocation rationale, independent agent risk review, intent router, and
the user-editable tax profile.}
\label{fig:dashboard}
\end{minipage}%
\hspace{0.25in}%
\begin{minipage}[t]{2.48in}
\centering
\includegraphics[height=\figH,valign=t]{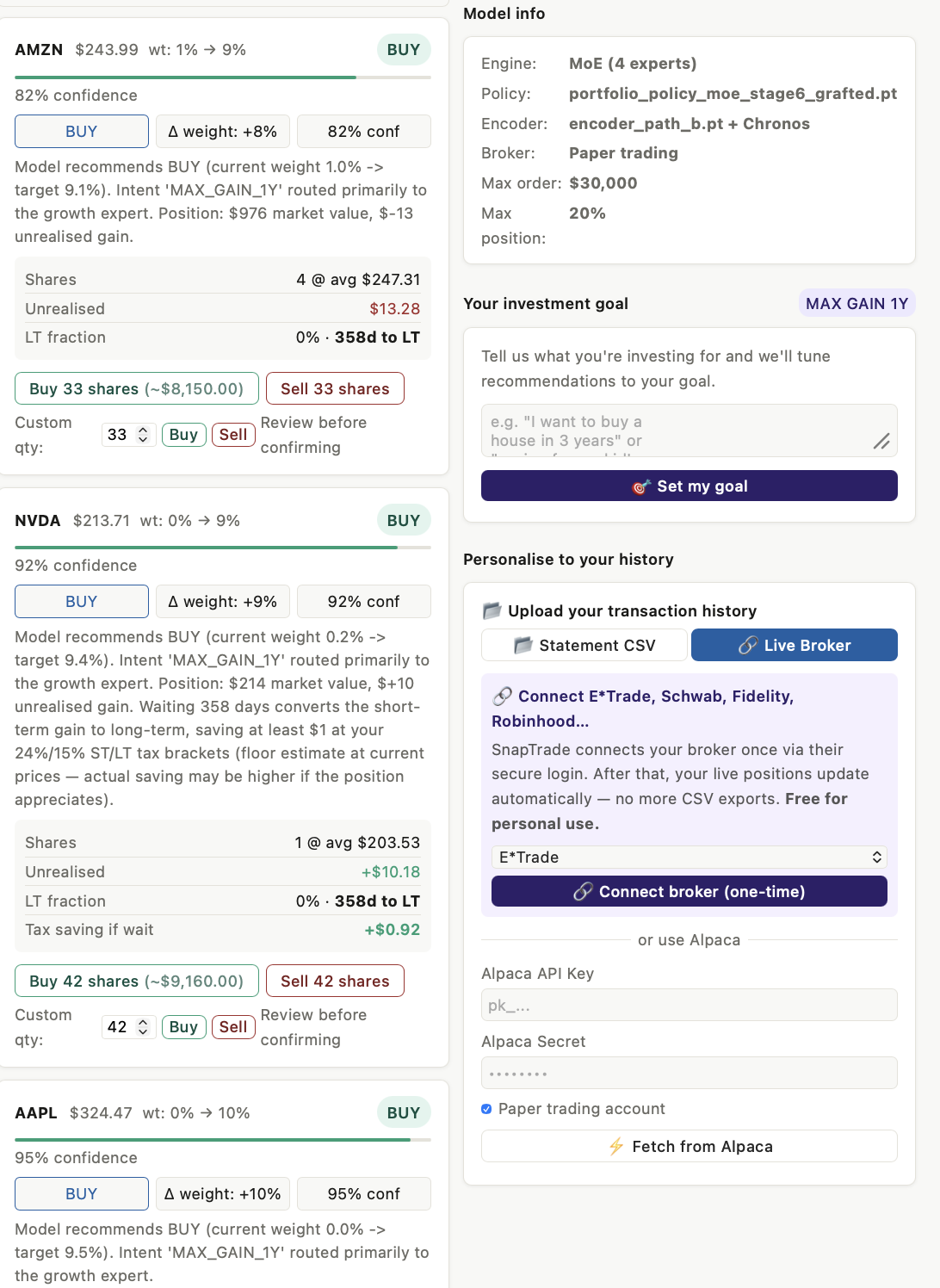}
\subcaption{Recommendation cards (action, weight change, confidence,
expert rationale, tax-lot reasoning, mandatory confirmation) and the
model/goal panel naming serving checkpoints, guardrails, and brokerage
connections.}
\label{fig:reccards}
\end{minipage}
\caption{The running dashboard. (a) portfolio and objective panels;
(b) per-ticker recommendations and model info.}
\label{fig:app}
\end{figure*}

The dashboard is a single-page web interface served directly by the
backend. Its main views are the current portfolio (positions, tax
lots, day and total P\&L from the connected broker); current
recommendations, each showing the proposed BUY/HOLD/SELL action,
target weights, the mandate and expert that drove it, and its tax
impact; a connect-brokerage panel (SnapTrade OAuth); and a
pending-confirmations queue implementing the preview-before-apply step
below. The intent is that a user never sees a bare model output: every
recommendation carries its rationale and consequences.
Figure~\ref{fig:app} shows the running application.

Figure~\ref{fig:dashboard} shows the portfolio overview and
objective/tax-profile panel. The header reports live integrity-chain
status (``Chain intact -- 27 entries (22 recommendations, 5 trades),
all externally timestamped''), surfacing the RFC~3161 audit trail
(Trust and Safety Controls) directly to the user rather than burying it
in a log. The left column renders the model's plain-language ``why this
allocation'' rationale alongside the agent risk review
(Trust and Safety Controls), so the reasoning and its audit appear
together. The intent router panel exposes the routing decision itself
-- here the free-text \texttt{MAX\_GAIN\_1Y} objective routed 100\% to
the growth expert, with the other three experts at zero -- making the
mixture weights visible rather than hidden inside the policy. The right
column is the user-editable tax profile (marginal income rate,
long-term capital-gains rate, horizon, annual taxable income) that
feeds the Phase~3 personalization layer, so a user can see and correct
the assumptions driving their own after-tax recommendations.

In the recommendation view
(Figure~\ref{fig:reccards}), each card states the proposed action, the
current-to-target weight change, a confidence score, and a rationale
naming the routed expert; the NVDA card surfaces tax-lot reasoning
(waiting 358 days converts the gain to long-term treatment), a
concrete instance of the tax-lot-aware behavior above. Every card carries an
explicit \emph{Review before confirming} step, realizing the
preview-before-apply guarantee, and the model-info panel names the
exact serving artifacts (grafted stage-6 MoE policy,
\texttt{encoder\_path\_b} plus Chronos), the active broker, and the
enforced guardrails.

\subsection{Natural Language Goal Specification}
A user's free-text goal is parsed into one of six mandate objectives
plus continuous risk/horizon parameters, replacing the
fixed-questionnaire pattern of existing robo-advisors. Parsing is
two-tier: a rule tier (keyword matching with regex year/month
extraction) handles common cases with no external dependency, and an
LLM tier parses the rest into a structured intent object (objective,
horizon, return target, drawdown tolerance, risk level) in under a
second. Table~\ref{tab:intents} shows representative mappings.

\begin{table*}[t]
\centering
\small
\caption{Example intent-to-objective mappings produced by the NL goal
parser.}
\label{tab:intents}
\resizebox{\textwidth}{!}{%
\begin{tabular}{lllr@{\hskip 2em}lllr}
\toprule
\textbf{Intent} & \textbf{Objective} & \textbf{Risk} & \textbf{Horizon} &
\textbf{Intent} & \textbf{Objective} & \textbf{Risk} & \textbf{Horizon} \\
\midrule
House in 3 years     & CAPITAL\_PRES. & Moderate     & 756d  &
Aggressive growth    & LT\_GAIN\_ONLY & Aggressive   & 1764d \\
College (kid age 10) & LT\_GAIN\_ONLY & Moderate     & 2016d &
Dividends, semi-ret. & INCOME\_HARV.  & Conservative & 504d  \\
Retiring in 2 years  & CAPITAL\_PRES. & Conservative & 504d  &
Emergency fund       & CAPITAL\_PRES. & Conservative & 252d  \\
Tax-loss harvesting  & INCOME\_HARV.  & Moderate     & 252d  &
                     &                &              &       \\
\bottomrule
\end{tabular}}
\end{table*}

\subsection{Tax-Lot-Aware Recommendations}
\label{sec:tax}
The Phase~3 personalization layer surfaces the tax consequence of a
proposed action directly in the recommendation -- for example
suppressing a SELL that is 26 days short of qualifying for long-term
capital-gains treatment, or flagging a tax-loss-harvesting
opportunity -- rather than reporting after-tax performance only as an
aggregate backtest statistic. Because the data layer stores true FIFO
tax lots rather than aggregate positions, these computations use the
user's actual acquisition dates and cost bases, so the holding-period
arithmetic shown to the user is their own rather than a modelled
approximation. Table~\ref{tab:tax_demo} shows the resulting behavior
for four investor personas on the same position, where the same model
gives materially different advice depending on lot age and bracket.

\subsection{Trust and Safety Controls}
\label{sec:safety}
We consider these controls the most important design decisions for a
system that makes financial recommendations autonomously.

\textbf{Preview-before-apply.} No recommendation reaches the brokerage
account without explicit user confirmation; the dashboard shows the
proposed change, rationale, and tax impact before anything executes,
converting every model output into a proposal a human reviews rather
than an action a model takes unilaterally.

\textbf{Hard pre-trade guardrails.} Configuration-enforced limits are
checked before any order is constructed: a maximum single-order value
(default \$5{,}000), a maximum single-ticker position fraction, a
daily-loss halt, and a redeployment-ratio cap on how fast idle cash is
deployed. These bound worst-case behavior independently of model
quality.

\textbf{Independent agentic risk review.} The central trust mechanism
is an LLM agent that reviews each recommendation downstream of the RL
policy, structurally separate from it, as a post-filter with
deliberately bounded authority: it can \emph{suppress} a proposed trade
(downgrade a \textsc{buy} or \textsc{sell} to \textsc{hold}) or
\emph{flag} it, but never introduce an action the policy did not
propose, reverse a direction, or relax a guardrail. It is genuinely
agentic in the tool-using sense of \citet{yao2023react} -- given three
real tools (recent news via yfinance and SEC EDGAR, earnings-event
lookups, and recent daily price action), it decides autonomously which
tickers and tools to investigate within a
bounded turn budget and emits a per-ticker verdict
(\textsc{suppress}, \textsc{hold\_flag}, \textsc{none}) with a written
reason shown to the user and recorded.

Two choices make delegating to an LLM defensible. First, the override
rules are \emph{hard numeric thresholds, not model judgment} (an
earnings event within two trading days, or a ten-day move
$\geq\!10\%$, forces a suppress; $6$--$10\%$ forces a flag) -- and the
deployed code does not trust the LLM to apply them: it captures the
real price-change figure the tool returned and
\emph{deterministically recomputes the band}, overriding the model's
stated category on any disagreement. The LLM decides what to
investigate; arithmetic decides the action. Second, the reviewer is
\emph{fail-safe and monotonic}: if it errors or is disabled it
produces no overrides and the pipeline is unchanged, and because it
can only restrict the action space, a reviewer failure can never widen
what reaches the user -- a scalable-oversight posture in the spirit of
\citet{amodei2016concrete}. The pre-trade guardrails
(above) are enforced separately at the broker layer,
beyond its reach.

Figure~\ref{fig:app}a shows this: the reviewer flags two positions on
price action alone (XOM $+12.21\%$, TSLA $-7.18\%$ over ten days) while
reporting that its earnings and news tools errored -- reasoning
transparently from what it obtained rather than proceeding silently. An
audit layer that announces its own blind spots and cannot
arithmetically outvote its own thresholds is, we argue, a precondition
for trusting an autonomous financial system.

\textbf{Auditable action-integrity chain.} Every recommendation and
trade is recorded in an append-only, per-user SHA-256 hash chain (each
entry links the previous entry's hash to the canonical-JSON hash of the
current payload) and is RFC~3161-timestamped at the moment it is
generated by an independent public timestamp authority
(freetsa.org). Because the timestamp is issued by a third party the
operator does not control, the log is tamper-evident: it proves what
the system recommended, what the reviewer said, and what the user
approved, and that each existed no later than its timestamp. Chains are
per-user rather than global, so one user's record is independently
verifiable and cannot be entangled with another's.

\textbf{Notifications.} A daily email digest delivers each user's
current recommendations, portfolio value, and pending confirmations,
so the confirmation queue does not silently go stale for users who do
not open the dashboard daily.
\begin{table*}[t]
\centering
\small
\caption{Phase~3 after-tax recommendations for four investor personas
on an illustrative fixed scenario (AAPL @ \$190.38, 15 Nov 2023); the
scenario is held fixed so the four personas differ only in lot age,
position, and bracket. ``LT saving'' = tax saving from waiting
for long-term treatment. ``sh'' is the number of shares held by a given investor. The system suppresses BUY/SELL actions that
would reduce after-tax value.}
\label{tab:tax_demo}
\resizebox{\textwidth}{!}{
\begin{tabular}{llllrr}
\toprule
\textbf{Persona} & \textbf{Objective} & \textbf{Bracket} &
\textbf{Action} & \textbf{After-tax now} & \textbf{Wait LT} \\
\midrule
30d trader, 20sh @ \$161.82
  & MAX\_GAIN\_30D & 32\%ST/15\%LT & HOLD & \$+388 & \$+485 \\
LT investor, 50sh @ \$133.26 (26d from LT)
  & LT\_GAIN\_ONLY & 24\%ST/15\%LT & HOLD$^\star$ & \$+2{,}170 & \$+2{,}427 \\
Loss position, 30sh @ \$237.97
  & INCOME\_HARVEST & 35\%ST/20\%LT & HOLD & \$-1{,}428 & \$-1{,}428 \\
Near-retirement, no position
  & CAPITAL\_PRESERVE & 22\%ST/15\%LT & HOLD & --- & --- \\
\bottomrule
\multicolumn{6}{l}{\small $^\star$Sell suppressed: 26d until LT conversion saves \$257 in tax.}
\end{tabular}}

\end{table*}

\section{Path to Full Deployment}
This application is functionally complete and integration-tested
against a live broker API, but has not been used by real end-users
with real capital, so we have no production usage data to report --
every number in this paper is pre-deployment validation, not evidence
of production performance.

Four concrete steps remain. First, a small pilot (order of 10--50
users) trading real but modest capital through the existing Alpaca
integration, instrumented to collect the usage data a Track~1
(Deployed Applications) submission would require: realised after-tax
returns against benchmarks, confirmation and override rates, how often
the risk reviewer suppresses or flags a trade and whether users agree,
and abandonment points in the goal flow. Second, a compliance review
of investment-adviser registration requirements, not yet completed and
which may restrict the pilot to paper trading or a small consenting
group until resolved. Third, broadening the validated ticker universe
beyond the ten equities evaluated here and moving to real-time market
data, since the encoder accepts an arbitrary universe at inference but
is only validated on this narrow one. Fourth, load-testing the
multi-user backend beyond the handful of accounts exercised so far,
including the inference and integrity-chain write paths under
concurrency.

None of these are architectural changes to the model -- they are the
deployment and validation work separating a tested application from a
deployed one, and are this project's natural next phase. The safety
architecture above was built \emph{before} rather than after this
pilot, which we regard as the correct ordering for a system with
brokerage write access.

\section{Preliminary Validation}
We report 14-day walk-forward backtests as \emph{pre-deployment}
validation of the underlying model -- evidence the recommendation
engine behind the application is sound -- not as production
performance, which does not yet exist. Table~\ref{tab:results} compares
four configurations on an identical window (10 tickers: AAPL, MSFT,
NVDA, AMZN, GOOGL, META, TSLA, JPM, XOM, V; equal-weight (EW) basket
return $-8.01\%$, SPY return $-2.76\%$).

\begin{table}[h]
\centering
\small
\caption{14-day walk-forward backtest, June 2026. Annualized Sharpe on
a 13-day window is uninformative and is reported, with caveats, in the
supplementary appendix.}
\label{tab:results}
\begin{tabular}{lcc}
\toprule
\textbf{Config} & \textbf{Alpha vs EW} & \textbf{Max DD} \\
\midrule
Single-head       & $+2.90\%$ & $-5.32\%$ \\
MoE (grafted)     & $+2.93\%$ & $-5.25\%$ \\
Chronos-only      & $+3.32\%$ & $-4.96\%$ \\
News+Chronos      & $+3.18\%$ & $-5.17\%$ \\
\bottomrule
\end{tabular}
\end{table}

All four configurations show positive alpha against the equal-weight
basket. Given the short window (13 daily returns), we quantify
uncertainty with a 10,000-resample day-level bootstrap: the
News+Chronos configuration's 95\% CI is $[-2.8\%, +9.6\%]$ ($84.9\%$ of
resamples positive) -- directionally encouraging but not significant
at conventional levels, and the differences between configurations are
well within this noise band. Alpha versus SPY is weaker
(approximately $-2\%$ across configurations), a structural consequence
of SPY's broader, more defensive constituent mix rather than a failure
of stock selection. We also note that the edge concentrates at short
horizons: in multi-window backtests the positive alpha persists
weakly to 30 days, all configurations are negative against
equal-weight at 60 days, and the grafted MoE recovers the best 90-day
result -- we report this pattern rather than selecting only the
favorable horizon. Full results, the multi-window table, the bootstrap
methodology, and an open-source script (\texttt{bootstrap\_ci.py}) for
reproducing this analysis are in the supplementary appendix and the
repository.

\section{Lessons Learned}
\label{sec:lessons}
Several practical lessons emerged while integrating live news/event
data, which we believe generalize beyond this application.

\textbf{Auto-detection covering one feature does not imply it covers a
structurally similar one.} Our pipeline auto-detected an optional
foundation-model encoder branch from checkpoint metadata; we assumed a
structurally similar news/event branch was covered by the same logic.
It was not -- the branch was saved and loadable but never invoked, for
two release cycles, because the detection logic was never extended to
it. Parallel optional features need independently verified activation,
not an assumption that a sibling code path works the same way.

\textbf{End-to-end empirical verification catches what code review
does not.} We discovered the above gap not through code review but
because live GPU utilization looked bursty in an unrelated debugging
session -- prompting an actual runtime trace (\texttt{py-spy}) rather
than a re-read of the code. For any system whose correctness depends on
an external data source actually being fetched and used, we recommend
a direct runtime check (e.g.\ instrumenting a call count, or a live
profiler snapshot) rather than trusting that a correctly-shaped
checkpoint or a clean code review implies the data path is live.

\textbf{Third-party APIs can hang, not just fail.} Two calls into a
financial-data library were found, via live profiling, to hang
indefinitely rather than raising -- one in an HTML-parsing fallback,
one in a stalled network call. \texttt{try/except} offers no protection
against a call that never returns, and a single hung request can occupy
a serving worker indefinitely. We now wrap every external API call in
a hard timeout (\texttt{signal.alarm}) and recommend this as a default,
not a reactive fix after a production stall.

\textbf{Caching must match the actual redundancy pattern.} A
per-request cache for earnings-date lookups gave no benefit, since the
API already returns a ticker's entire history per call; caching
per-ticker instead of per-query cut a full refresh pass from an
estimated tens of thousands of calls to ten -- profile what the
upstream API returns before choosing a cache key.

\textbf{Model-serving dependencies dominate deployment friction.} Early
deployments failed not on model code but on packaging: the default
CUDA-bundled PyTorch distribution exceeded the hosting platform's
build-image budget with gigabytes the CPU serving path never uses.
Pinning the CPU-only wheel fixed both build failures and cold-start
time. For small policies served on CPU, treat the inference dependency
set as a deliberately minimal artifact, separate from the training
environment, from day one.

\section{Reproducibility and Code Availability}
The full system -- training pipeline, inference engine, FastAPI
backend, dashboard, brokerage abstraction, risk-review agent, and
integrity chain -- is open source at
\url{https://github.com/rpishehvar/PublicFinance-RL}, with the
database schema needed for a multi-user instance. Training scripts fix
a global seed (Python, NumPy, PyTorch); cuDNN determinism flags are
not set, so bit-exact cross-GPU reproduction isn't guaranteed, though
reported metrics were stable across repeated runs. Package versions
are pinned in the repository, and the supplementary appendix records
training hardware, wall-clock cost, and the full PPO configuration.

Two caveats. Backtests depend on third-party market data whose
historical values can be revised, so exact replication needs the
cached data snapshots rather than a fresh fetch; the bootstrap script
producing our confidence intervals is included so uncertainty can be
regenerated from the same returns. The risk-review agent calls a
hosted LLM, so its verdicts aren't bit-reproducible across runs -- which
is why override thresholds are enforced deterministically outside the
model, and why disabling the agent leaves the rest of the pipeline
unchanged and reproducible.

\section{Related Work}
Existing retail robo-advisors allocate accounts using static,
rule-based glide paths tied to age and a coarse risk questionnaire
\citep{aacunto2019,beketov2018robo}, which our system replaces with a
learned policy conditioned on a free-text goal. Prior applied RL
portfolio work \cite{jiang2017,liu2021,ye2020} learns on historical
data but targets simulated or backtested performance rather than a
broker-integrated, user-facing application with natural-language goal
input and a trust-first confirmation flow.

\section{Conclusion}
We presented a functionally complete, integration-tested retail
portfolio management application -- live broker integration,
multi-user support, natural-language goals, tax-lot-aware
recommendations, and a trust-first confirmation UX governed by an
independent agentic risk reviewer -- built on a three-phase RL system
and reported honestly as \emph{emerging} rather than deployed.
Preliminary backtests show a real but statistically modest edge over
equal-weight, with confidence intervals reported rather than
suppressed. We believe the safety architecture -- deterministic
guardrails no language model can relax, an audit layer that announces
its own blind spots, and a tamper-evident record of every
recommendation -- and the engineering lessons from integrating live
financial data are of independent value to applied RL practitioners.

\newpage

\onecolumn
\begin{center}
{\Large\bfseries Supplementary Technical Appendix:\\[2pt]
An Emerging Retail Portfolio Management Application\par}
\vspace{6pt}
{\normalsize Ramin Pishehvar \quad \texttt{rpichevar@gmail.com}\par}
\vspace{10pt}
\end{center}
\small
\section{Appendix: Model, Training, and Extended Results}
This appendix is supplementary. Full derivations, ablations, and diagnostics beyond
what fits here are in the arXiv version of the underlying model
paper~\cite{pishehvar2026threephase} and the open-source repository.

\subsection{Model and Training Details}
\label{app:model}

\paragraph{Phase 1: self-supervised pretraining.}
The encoder is pretrained on a 30-ticker S\&P~500 sample (all 11 GICS
sectors, daily bars 2015--2024) with four objectives -- next-bar
return prediction (Huber, $\delta=0.05$) \cite{sun2018adaptive},
masked feature recovery, four-way market-regime classification, and an
inter-ticker contrastive term:
\begin{equation}
\mathcal{L}_{P1} = 0.3\mathcal{L}_\text{ret} + 1.0\mathcal{L}_\text{mask}
 + 0.5\mathcal{L}_\text{reg} + 0.5\mathcal{L}_\text{contrast},
\qquad
\mathcal{L}_\text{contrast} = \frac{1}{BN(N{-}1)}
\sum_{b=1}^{B}\sum_{i\neq j}
\frac{h_{b,i}\cdot h_{b,j}}{\lVert h_{b,i}\rVert\,\lVert h_{b,j}\rVert}.
\end{equation}
The contrastive term exists because standard SSL objectives treat each
ticker independently and provide no differentiation signal: without
it, cross-asset attention converged to a mean representation (mean
inter-ticker cosine similarity 0.96) and the allocation head output
uniform $1/N$ weights regardless of input. With
$\lambda_\text{contrast}=0.5$, similarity drops to 0.24 over 60 epochs
and allocation weights become genuinely differentiated
(Table~\ref{tab:phase1}).

\paragraph{Chronos fusion.}
A frozen Chronos-T5 time-series foundation model provides a parallel
embedding of each ticker's raw closing-price sequence, combined with
the SSL representation via a learned gate, with 50-dimensional
observable metadata (sector one-hot, market-cap bucket, fundamentals,
analyst consensus, options signals, earnings calendar, technical
regime, insider and institutional activity) added afterwards:
\begin{equation}
\tilde h_i = h^{\text{ssl}}_i
 + \sigma\!\big(W_g[h^{\text{ssl}}_i; h^{\text{chr}}_i]\big)\odot h^{\text{chr}}_i
 + \text{MetadataEnc}(m_i).
\end{equation}
Only the projection (74K params) and gate (8K params) are trained;
99.97\% of Chronos weights stay frozen, and embeddings for the
training corpus are precomputed once and cached. Because metadata is
observable for any ticker, the model has no ticker-identity embedding
table and accepts any universe size $N$ at inference.

\paragraph{Phase 2: objective-conditioned portfolio PPO.}
A portfolio actor-critic is fine-tuned with PPO
\cite{schulman2017proximal}, jointly outputting allocation weights
$w\in\Delta^{N+1}$ (softmax over $N$ equities plus a learnable cash
token, forcing an active cash decision every step) and per-ticker
HOLD/BUY/SELL actions. The environment rebalances whenever
$|w^{\text{eq}}_i - w^{\text{curr}}_i| > \delta_\text{reb}$ (default
0.01), with the action head acting as a modifier rather than a gate --
a decoupling required after the action head collapsed to
unconditional HOLD in early experiments. Each episode samples one of
six objectives $o$, shaping the reward
\begin{equation}
R_o = S_o
 - \lambda_c H_\text{eq}
 - \lambda_t \tilde\tau
 - \gamma_d \max(0, c_r - 0.05)
 - \gamma_s\,\mathbf{1}[\tau{=}0,\, t{>}10]
 + \delta_r\,\rho,
\end{equation}
where $S_o$ is the objective-specific base score (Sharpe-based for
growth mandates, drawdown-penalized for capital preservation,
holding-period- and harvest-aware for the tax mandates), $H_\text{eq}$
is the equity-only Herfindahl concentration \cite{hirschman1945national},
$\tilde\tau$ is turnover net of same-step sell$\to$buy round-trips,
$c_r$ the cash ratio, and $\rho$ the fraction of sell proceeds
redeployed in the same step. A hard per-step turnover cap
($\tau_\text{cap}=0.25$) and an action-entropy floor prevent the
degenerate high-turnover and all-HOLD solutions respectively.

\paragraph{MoE curriculum and expert grafting.}
Training a single head on all six objectives simultaneously collapses
to uniform $1/N$ weights within 100 episodes from gradient conflict.
We instead train four expert heads (momentum, growth, defensive,
tax-aware; objectives partitioned across them) through the six-stage
curriculum in Table~\ref{tab:curriculum}. Router-only training with
frozen experts (Stage~4) fails on a flat loss surface; Stage~5
resolves it with an intent-projection layer, a diagonal
intent$\to$expert shortcut initialization, and a supervised routing
loss ($\lambda_\text{sup}=1.0$, router LR scaled $10\times$),
achieving clean one-hot routing by episode~33. Because joint training
degrades specialist quality (the momentum expert's 14d alpha falls
from $+3.37\%$ to $+0.01\%$), Stage~6 \emph{grafts} the best
per-expert curriculum checkpoints under the Stage~5 router with no
further training, recovering $+3.03\%$ 14d alpha and the best 90d
result ($-2.11\%$, Table~\ref{tab:multiwindow}).

\begin{table}[!ht]
\centering
\small
\caption{MoE expert curriculum. All stages: $N=10$ tickers,
$\beta_H=0.02$, $\lambda_\tau=0.05$, $\tau_\text{cap}=0.25$,
$\delta_\text{reb}=0.01$. Each of Stages~1--3 trains one expert
exclusively while the others are frozen.}
\label{tab:curriculum}
\resizebox{\textwidth}{!}{%
\begin{tabular}{lllcll}
\toprule
\textbf{Stage} & \textbf{Expert / goal} & \textbf{Objective} &
\textbf{Ep.} & \textbf{Frozen experts} & \textbf{Enc.\ unfreeze} \\
\midrule
1 & Momentum expert  & ALPHA\_VS\_EW      & 300 & 1,2,3   & ep 50 \\
2 & Growth expert    & MAX\_GAIN\_1Y      & 300 & 0,2,3   & ep 50 \\
3 & Defensive expert & CAPITAL\_PRESERVE  & 300 & 0,1,3   & ep 50 \\
4 & Router-only      & All (experts frozen) & 300 & 0,1,2,3 & never \\
5 & Joint + supervised router & All       & 300 & none    & ep 50 \\
6 & Grafted MoE      & S2/S3 experts + S5 router & --- & --- & --- \\
\bottomrule
\end{tabular}}
\end{table}

\paragraph{Phase 3: LoRA personalization.}
A tax-aware personalization layer adapts only action logits via a
low-rank adapter \cite{hu2021lora}:
$\hat\ell = \ell_\text{base} + p_u A B$, where $p_u\in\mathbb{R}^{16}$
is a behavior profile extracted from brokerage transaction history
(median holding period, LT-sell fraction, loss-harvest score,
disposition effect, trade frequency) and $A\in\mathbb{R}^{16\times r}$,
$B\in\mathbb{R}^{r\times 3}$ with $r=4$, $B$ initialized to zero so
adaptation is an identity at deployment. Total adapter size: 76
parameters ($\approx$1\,KB) persisted per user; encoder and experts
stay frozen.

\subsection{Extended Results}
\label{app:results}

\begin{table}[!ht]
\centering
\small
\caption{Phase~1 ablation: validation loss and mean inter-ticker
cosine similarity (60 epochs each). Chronos' input normalization
causes representation collapse (sim 0.81--0.96) that the contrastive
loss corrects; warm-starting the contrastive phase from SSL-pretrained
weights is best.}
\label{tab:phase1}
\begin{tabular}{lccl}
\toprule
\textbf{Configuration} & \textbf{Val loss} & \textbf{Sim.} & \textbf{Note} \\
\midrule
Path B, no Chronos            & 0.277 & 0.08 & raw features differentiate \\
Path B, Chronos-tiny          & 0.171 & 0.81 & 8M frozen params \\
Path B, Chronos-small         & 0.171 & 0.81 & 46M frozen, no contrastive \\
+ Contrastive, scratch        & 0.305 & 0.11 & from random init \\
+ Contrastive, warm-start     & 0.163 & 0.24 & SSL-pretrained init \\
\bottomrule
\end{tabular}
\end{table}

\begin{table}[!ht]
\centering
\small
\caption{Full 14-day walk-forward backtest, June 2026. 10 tickers,
\$100{,}000 initial capital, zero transaction cost. The four
right-hand columns share an identical window (EW $-8.01\%$, SPY
$-2.76\%$); the collapsed-representation column is an earlier
diagnostic run on a separate window (own EW $-5.21\%$) included to
illustrate the collapse failure mode. Chronos-only and News+Chronos
are sequential-specialist checkpoints (no joint fine-tuning or
grafting).}
\label{tab:fullbacktest}
\resizebox{\textwidth}{!}{%
\begin{tabular}{lrrrrr}
\toprule
\textbf{Metric} & \textbf{Collapsed} & \textbf{Single-head} &
\textbf{MoE (grafted)} & \textbf{Chronos-only} & \textbf{News+Chronos} \\
\midrule
Total return      & $-5.95\%$ & $-5.11\%$ & $-5.07\%$ & $-4.68\%$ & $-4.83\%$ \\
Alpha vs EW       & $-0.74\%$ & $+2.90\%$ & $+2.93\%$ & $+3.32\%$ & $+3.18\%$ \\
Alpha vs SPY      & $-3.53\%$ & $-2.35\%$ & $-2.31\%$ & $-1.92\%$ & $-2.07\%$ \\
Ann.\ Sharpe      & $-6.91$   & $-6.47$   & $-6.62$   & $-6.51$   & $-6.30$ \\
Max drawdown      & $-5.87\%$ & $-5.32\%$ & $-5.25\%$ & $-4.96\%$ & $-5.17\%$ \\
Win rate (daily)  & 35.7\%    & 35.7\%    & 42.9\%    & 42.9\%    & 35.7\% \\
\bottomrule
\end{tabular}}
\end{table}

\paragraph{News-branch correction.}
The run originally labeled "news-fused" in an earlier draft is retained above under its correct label (Chronos-only, +3.32\%), since it did not include an active news branch. With the news branch genuinely active end-to-end, the retrained News+Chronos checkpoint achieves +3.18\% — a positive point estimate, but not statistically distinguishable from the no-news baseline under the bootstrap below, so we claim stable training but not a demonstrated improvement from news fusion.

\paragraph{Statistical uncertainty.}
All point estimates come from a single 14-trading-day window (13 daily
returns). A 10{,}000-resample day-level bootstrap (resampling daily
returns with replacement, compounding per resample, benchmark return
held at its realized value) gives 95\% CIs of $[-2.3\%, +9.2\%]$ for
Chronos-only and $[-2.8\%, +9.6\%]$ for News+Chronos -- both include
zero, with 87.6\% and 84.9\% of resamples positive respectively.
We therefore read Table~\ref{tab:fullbacktest} as consistent with a
real but modest edge over equal-weight for the Chronos-augmented
family as a whole, not as a reliable ordering between configurations.
Alpha vs SPY is negative ($\approx -2\%$) for all configurations, a
structural consequence of the growth-heavy 10-ticker universe versus
SPY's broader defensive mix during this window; the negative
annualized Sharpe values likewise reflect a declining window
(EW $-8.01\%$) rather than systematic underperformance -- on short
windows, alpha vs the same-universe equal-weight basket is the metric
that isolates selection skill. The analysis script
(\texttt{bootstrap\_ci.py}) is included in the repository.

\begin{table}[!ht]
\centering
\small
\caption{Best-epoch alpha vs equal-weight (EW) and SPY across backtest
windows and curriculum stages. Window-adaptive rebalancing thresholds:
1.0 (14d), 0.05 (30d), 0.03 (60d), 0.02 (90d); zero transaction cost.
S4 (router-only, frozen experts) fails on a flat loss surface.}
\label{tab:multiwindow}
\resizebox{\textwidth}{!}{%
\begin{tabular}{lrrrrrrrr}
\toprule
\textbf{Stage} &
  \multicolumn{2}{c}{\textbf{14d}} &
  \multicolumn{2}{c}{\textbf{30d}} &
  \multicolumn{2}{c}{\textbf{60d}} &
  \multicolumn{2}{c}{\textbf{90d}} \\
\cmidrule(lr){2-3}\cmidrule(lr){4-5}\cmidrule(lr){6-7}\cmidrule(lr){8-9}
 & EW & SPY & EW & SPY & EW & SPY & EW & SPY \\
\midrule
S1: Momentum (ep100)
 & $+3.16\%$ & $-2.08\%$ & $+0.21\%$ & $-7.24\%$ & $-3.29\%$ & $-11.33\%$ & $-2.10\%$ & $-5.44\%$ \\
S2: Growth (ep100)
 & $+3.37\%$ & $-1.88\%$ & $+0.19\%$ & $-7.27\%$ & $-4.12\%$ & $-12.16\%$ & $-2.37\%$ & $-5.70\%$ \\
S3: Defensive (ep100)
 & $+3.18\%$ & $-2.07\%$ & $+0.11\%$ & $-7.35\%$ & $-4.07\%$ & $-12.12\%$ & $-2.42\%$ & $-5.76\%$ \\
S4: Router-only (ep300)
 & $+0.44\%$ & $-3.43\%$ & --- & --- & --- & --- & --- & --- \\
S6: Grafted MoE
 & $+3.03\%$ & $-4.07\%$ & $+0.17\%$ & $-6.05\%$ & $-4.20\%$ & $-12.24\%$ & $-2.11\%$ & $-5.97\%$ \\
\bottomrule
\end{tabular}}
\end{table}

\paragraph{Multi-window pattern.}
The 14-day positive alpha does not persist at longer horizons
(Table~\ref{tab:multiwindow}): momentum and growth experts hold a
small positive edge through 30 days, all configurations are negative
at 60 days, and the grafted MoE achieves the best 90-day result --
the per-horizon specialization the MoE router is designed to exploit.
At Stage~5, all six intents route one-hot to their designated experts
(routing std $>0.37$ per expert against a 0.05 threshold) with
distinct top holdings per expert (JPM momentum, TSLA growth, GOOGL
defensive, AMZN tax-aware), confirming meaningfully different learned
strategies.

\subsection{Reproducibility}
\label{app:repro}
All training scripts fix \texttt{SEED = 42} (Python, NumPy, PyTorch);
cuDNN determinism flags were not set, so bit-exact cross-GPU
reproduction is not guaranteed, though reported metrics were stable
across repeated runs. Training ran on a single NVIDIA L4 (24\,GB),
PyTorch 2.8.0 (CUDA 12.8), Python 3.12, with Chronos embeddings from
\texttt{amazon/chronos-t5-small} via \texttt{chronos-forecasting};
exact versions are pinned in the repository's
\texttt{requirements.txt}. One MoE curriculum stage (300 PPO episodes,
rollout length 512, 10 tickers) takes $\sim$25--30 minutes
($\sim$5\,s/episode); total wall-clock for Phase~1 and the full
curriculum was not separately profiled. Table~\ref{tab:ppo} lists the
PPO configuration; the entropy coefficient is raised per-stage where
noted (0.05 for the momentum expert).

\begin{table}[!ht]
\centering
\small
\caption{PPO hyperparameters (Adam, $\epsilon_\text{Adam}=10^{-5}$),
Phase~2 and MoE curriculum.}
\label{tab:ppo}
\begin{tabular}{lr}
\toprule
\textbf{Hyperparameter} & \textbf{Value} \\
\midrule
Discount factor $\gamma$        & 0.99 \\
GAE $\lambda$                   & 0.95 \\
PPO clip $\epsilon$             & 0.2 \\
Value loss coefficient          & 0.5 \\
Entropy coefficient (default)   & 0.01 \\
Gradient norm clip              & 0.5 \\
Learning rate (default / ALPHA\_VS\_EW) & $10^{-4}$ / $10^{-3}$ \\
PPO epochs / minibatch          & 4 / 32 \\
Rollout length                  & 256--512 \\
\bottomrule
\end{tabular}
\end{table}

\end{document}